\documentclass[letterpaper, 10 pt, conference]{ieeeconf}
\IEEEoverridecommandlockouts
\usepackage{cite}
\usepackage{amsmath,amssymb,amsfonts}
\usepackage{algorithm}
\usepackage[noend]{algpseudocode}
\usepackage{graphicx}
\usepackage{textcomp}
\usepackage{xcolor}
\usepackage{multirow}
\usepackage{siunitx}
\usepackage{adjustbox}
\usepackage{array}
\usepackage{tikz}
\usepackage{pgfplots}
\usepackage{caption}
\usepackage{hyperref}
\usepackage{dsfont}

\pgfplotsset{compat=1.18}
\usepackage[font=small]{caption}

\pgfplotsset{
every axis/.append style={
  axis line style={->}, 
  legend style={font=\footnotesize},
  label style={font=\footnotesize},
  title style={font=\footnotesize},
  tick label style={font=\footnotesize},
  axis x line*=bottom,
  axis y line*=left,
  }
}

\def\BibTeX{{\rm B\kern-.05em{\sc i\kern-.025em b}\kern-.08em
    T\kern-.1667em\lower.7ex\hbox{E}\kern-.125emX}}
\begin{document}
\title{\bf \LARGE RISeg: Robot Interactive Object Segmentation via\\ Body Frame-Invariant Features}

\author{Howard H. Qian$^{1}$, Yangxiao Lu$^{2}$, Kejia Ren$^{1}$, Gaotian Wang$^{1}$, Ninad Khargonkar$^{2}$, Yu Xiang$^{2}$, Kaiyu Hang$^{1}$
\thanks{$^{1}$Department of Computer Science, Rice University, Houston, TX 77005, USA. HQ, KR, GW, and KH are supported by the US National Science Foundation grant FRR-2133110. $^{2}$Department of Computer Science, University of Texas at Dallas, Richardson, TX 75080, USA. YL, NK and YX are supported by the DARPA Perceptually-enabled Task Guidance (PTG) Program under contract number HR00112220005 and the Sony Research Award Program.}
}

\maketitle

\begin{abstract}
In order to successfully perform manipulation tasks in new environments, such as grasping, robots must be proficient in segmenting unseen objects from the background and/or other objects. Previous works perform unseen object instance segmentation (UOIS) by training deep neural networks on large-scale data to learn RGB/RGB-D feature embeddings, where cluttered environments often result in inaccurate segmentations. We build upon these methods and introduce a novel approach to correct inaccurate segmentation, such as under-segmentation, of static image-based UOIS masks by using robot interaction and a designed body frame-invariant feature. We demonstrate that the relative linear and rotational velocities of frames randomly attached to rigid bodies due to robot interactions can be used to identify objects and accumulate corrected object-level segmentation masks. By introducing motion to regions of segmentation uncertainty, we are able to drastically improve segmentation accuracy in an uncertainty-driven manner with minimal, non-disruptive interactions (\emph{ca.} 2-3 per scene). We demonstrate the effectiveness of our proposed interactive perception pipeline in accurately segmenting cluttered scenes by achieving an average object segmentation accuracy rate of 80.7\%, an increase of 28.2\% when compared with other state-of-the-art UOIS methods.
\end{abstract}

\section{Introduction}

In order to perform autonomous manipulation tasks, robots must be able to robustly perceive and segment unseen objects to gain an understanding of their environment. Thus, competent unseen object instance segmentation (UOIS) is imperative to a robot's manipulation capabilities\cite{Xiang2021UOIS,Back2022Amodal,Xie2020UOIS,Danielczuk2018}.

While many state-of-the-art UOIS methods leverage deep neural networks to extract pixel-wise feature representations to perform segmentation, under and over segmentation in cluttered scenes remain a challenge \cite{Xiang2021UOIS, MSMFormer}. Because these methods attempt to segment single RGB-D images, only visual features are modeled while some essential physical features, such as how adjacent objects move relatively to one another, are not considered. Interactive perception is an alternative UOIS approach in which robots physically interact with the environment to accumulate information over time \cite{Brock2009}. Under interactive perception, we should aim to gather the most sensory data from interactions with as little amount of scene disturbance as possible. For example, if our robot's main task is to clean wine glasses, we must first identify the wine glasses by segmenting them out from the background. While interactively segmenting the scene, we should minimize our physical disturbances as to not accidentally knock over and break the glasses.

Central to the proposed method is our designed body frame-invariant feature (BFIF). Assuming there are two body frames rigidly attached to an object. We build our system on the insight that, when this object is moving, although the two body frames are rotating and translating differently in space, they will have the same spatial twist as observed by any reference frame fixed to the world\cite{ModernRobotics}. This fact applies to arbitrarily many body frames. Meanwhile, body frames on different objects that are relatively moving will typically have different spatial twists. This insight enables the design of BFIF for robot interaction-based object segmentation.

\begin{figure}[t]
\includegraphics[width=\columnwidth]{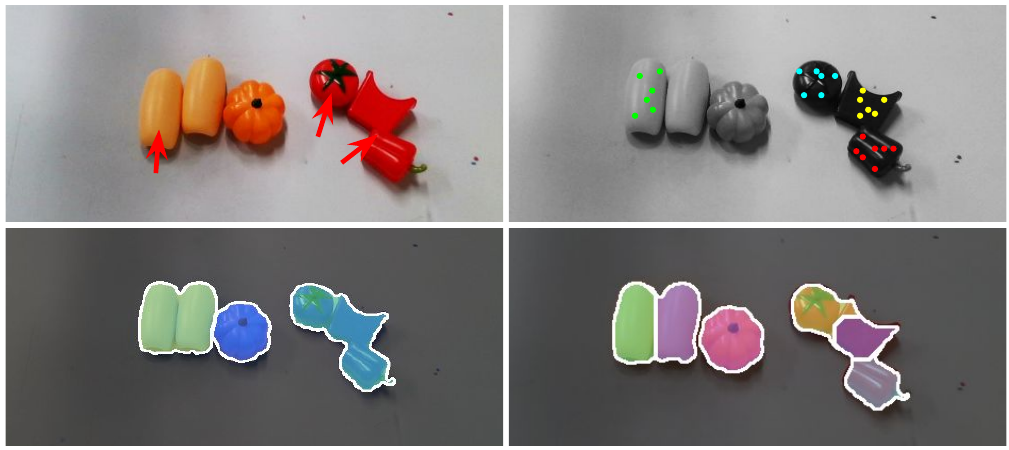}
\centering
\caption{Interactively segmenting a cluttered scene with minimal, non-disruptive pushes. [Top left] Initial scene and identified robot actions. [Top right] The origins of sampled body frames with matched BFIFs due to scene interactions, where matched body frames share the same color. [Bottom left] Undersegmentation of scene's end configuration by static segmentation model. [Bottom right] Accurate segmentation of scene by RISeg after interactions have been completed.}
\label{mainFig}
\vspace{-20pt}
\end{figure}

This work proposes the framework of Robot Interactive Segmentation (RISeg), which leverages active robot-object interactions and the BFIF to improve the performance of UOIS. Rather than learning visual features via data\cite{Xiang2021UOIS}, we demonstrate that segmentation of complex, cluttered scenes can be drastically improved by observing object motions and grouping BFIFs throughout robot interactions (see Fig.~\ref{mainFig}). Singulation of objects at any step of robot interaction is not necessary for our method, which results in fewer pushes (\emph{ca.} 2-3) and less disturbance to environments when compared to prior interactive perception methods \cite{lu2023selfsupervised}. 

\vspace{-5pt}
\section{Related Work}

\subsection{Unseen Object Instance Segmentation}

Unseen object instance segmentation is the task of segmenting all object instances within an image without prior knowledge about said objects \cite{Xiang2021UOIS}. Early works in UOIS utilize low-level image features such as edges, contours, or convexity to group pixels with one another \cite{Felzenszwalb2004Graph, Trevor2013EfficientOP, Stein2014Convexity, Pham2017SceneCut, forsyth2003computer}. Since these methods consider all such details within an image without an object-level understanding, objects are often over-segmented. More recent works make use of deep neural networks 
and large-scale training data, which has led to significantly better performance \cite{Xie2019Both, Danielczuk2019Data, Shao2018ClusterNetIS, kirillov2023segmentanything}. However, the challenges of bridging the sim-to-real gap, avoiding training data biases, or overcoming object-to-object occlusions tend to result in under-segmentation of real images \cite{Zhang2023UOIS,Balloch2018Synthetic}. While both low-level and learning-based methods only use single images and are limited in real-world performance, we show that, without requiring any changes to a learned segmentation neural network, the proposed RISeg will dramatically improve its real-world performance, especially upon under-segmentation failure cases.

\subsection{Motion-Based Robot Perception}

Motion-based segmentation methods attempt to segment environments by utilizing a robot's interactions with objects to detect scene changes in a sequence of images \cite{Brock2009, Bohg2016}. Previous works in this field fall under various categories such as statistical, factorization, or image differencing methods \cite{zappella2008motion, Costeira1998AMF, Goh2007, Arsenio, Fitzpatrick2003, metta2003early}. These methods, however, either require prior knowledge of objects, are computationally expensive, or can only segment objects that have been moved. Furthermore, multi-view scene perception methods utilize images captured from different viewpoints and segment objects based on consistencies across changing views \cite{Zeng2017MultiView,Mitash2017MultiView}. Yet, these methods often see similar failure cases as methods which use single images due to lack of object-level motions \cite{Brock2009}. Another type of motion-based robot perception approach utilizes video motion-tracking throughout an action to segment objects \cite{lu2023selfsupervised}. While these methods similarly only segment objects that have been interacted with or require object singulation over a long sequence of actions, our method is able to segment objects using a minimal number of non-disruptive actions by interacting with objects close to identified regions of uncertainty.

\section{Problem Formulation}

In this section, we will formally define the interactive perception problem and introduce our proposed method. Our system breaks down the interactive perception framework of ``observe, interact, observe'' into 2 main contributions in action planning and segmentation mask correction.

As previously mentioned, we should aim to maximize our understanding of a given environment while minimizing scene disturbance throughout interactions. By using segmentation masks predicted by a static image-based model before and after each interaction, we are able to make interaction decisions and improve object segmentations based on the scene's motions.

To formalize our proposed method, let $I_t \in [0, 255]^{H \times W \times 3} \times \mathbb{R}_{+}^{H \times W}$ be the RGB-D image of the given scene at time step $t$, where $t = 0, 1, 2, \ldots$, is the discrete time of the system. Let the inputs to our interactive perception system, RISeg, be an RGB-D image of the scene's initial state, $I_{0}$, and a static RGB-D image segmentation model $\Call{StaticSeg}{\cdot}$. The model $\Call{StaticSeg}{\cdot}$ takes image $I_t$ as an input and outputs a segmentation mask $L_t \in \mathbb{Z}_{+}^{H \times W}$. $L_t^{i,j} \in L_t$ indicates a pixel-wise object ID of pixel $(i,j)$ in $I_t$. If $L_t^{i,j} = 0$, then pixel $(i,j)$ of $I_t$ is segmented as part of the background. For all other integer values $L_t^{i,j} > 0$, pixel $(i,j)$ of $I_t$ is predicted to be part of an object. For example, $L_t^{i,j} = 1$ indicates that pixel $(i,j)$ of $I_t$ is part of object $1$.

In Alg.~\ref{RISeg}, we algorithmically describe a system in which the scene is observed between interactions to produce more accurate segmentation masks. After each interaction, $a_t \in SE(3)$, is identified by $\Call{FindAction}{\cdot}$ and completed by $\Call{Interact}{\cdot}$, a segmentation mask, $\hat{L}_{t+1} \in \mathbb{Z}^{H \times W}_+$, is produced by $\Call{UpdateMask}{\cdot}$ through BFIF analysis. Once the stop condition is met, the final segmentation mask $\hat{L}_{t+1}$ is returned which reflects a more accurate segmentation of the scene's end configuration after interactions.

\begin{figure}
\vspace{-5pt}
\begin{algorithm}[H]
\footnotesize
\caption{RISeg}\label{RISeg}
\textbf{Input:} $I_{0}$, $\Call{StaticSeg}{\cdot}$ \\
\textbf{Output:} $\hat{L}_{t+1}$
\begin{algorithmic}[1]
\State $t \leftarrow 0$
\State $L_{t} \leftarrow \Call{StaticSeg}{I_t}$
\State $\hat{L}_{t} \leftarrow L_{t}$
\While {$a_{t} \leftarrow \Call{FindAction}{I_{t}}$ \textbf{not} \textit{null}} \Comment{Alg.~\ref{FindAction}}
\State $I_{t+1} \leftarrow \Call{Interact}{a_{t}}$
\State $L_{t+1} \leftarrow \Call{StaticSeg}{I_{t+1}}$
\State $\hat{L}_{t+1} \leftarrow \Call{UpdateMask}{I_{t}, I_{t+1}, \hat{L}_{t}, L_{t+1}}$ \Comment{Alg.~\ref{UpdateMask}}
\State $t \leftarrow t+1$
\EndWhile
\State \textbf{return} $\hat{L}_{t+1}$
\end{algorithmic}
\end{algorithm}
\vspace{-20pt}
\end{figure}

\section{Body Frame-Invariant Feature}

\begin{figure}[htbp]
\includegraphics[width=\columnwidth]{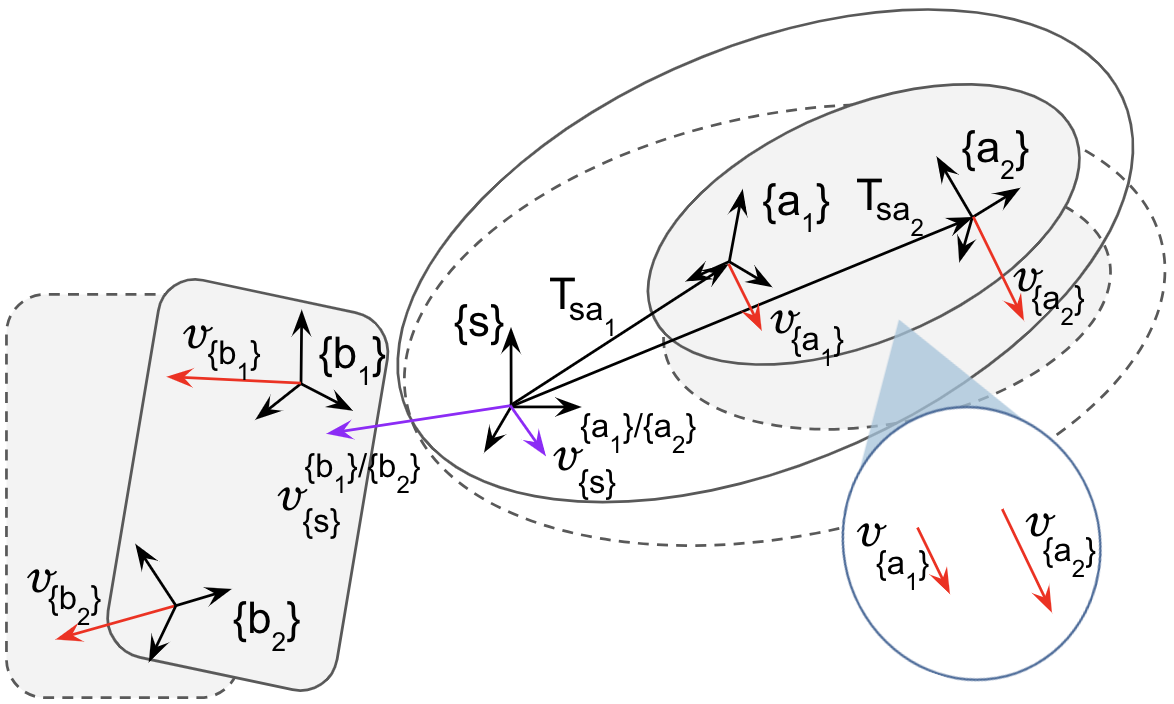}
\centering
\caption{A visual representation of BFIFs. Motions of different body frames attached to the same rigid body are transformed into the same space frame twist. Sampled body frames $\{a_1\}$ and $\{a_2\}$ lie on the shaded oval object and $\{b_1\}$ and $\{b_2\}$ lie on the shaded rectangle object. Space frame $\{s\}$ is arbitrarily chosen. Body frames are shown on the initial (solid line) configurations of the rigid bodies and corresponding motions onto the displaced (dashed line) rigid body configurations are represented by linear velocity vectors $\upsilon_{\{x\}}$ (red). The closeup circle shows $\upsilon_{\{a_1\}} \neq \upsilon_{\{a_2\}}$. Transparent oval shapes show the shaded oval object imagined to be infinitely large. Linear velocities of each body frame $\upsilon_{\{x\}}$ are transformed to the space frame and are shown by spatial velocity vectors (purple). Corresponding body frames for each spatial velocity vector are denoted in the superscript of $\upsilon_{\{s\}}$.}
\label{SampledFrames}
\vspace{-10pt}
\end{figure}

The proposed RISeg method is an interactive perception method in which a designed body frame-invariant feature (BFIF) of sampled frames within a scene are grouped with one another based on computed feature similarities. BFIF is based on the spatial twists of body frames attached to various rigid bodies. The key point being that twists of moving body frames on the same rigid body transformed into a fixed space frame will all have the same spatial twist, no matter their relative motion (see Fig.~\ref{SampledFrames}) \cite{ModernRobotics}. A frame defines a coordinate system with $X$, $Y$, and $Z$ axes attached to an origin in $SE(3)$. 

Given a body frame $\{b\}$ attached to a rigid body that experiences some translation and/or rotation, the motion of $\{b\}$ can be derived as a twist $\mathcal{V}_b$. The body twist $\mathcal{V}_b$ represented in the $\{b\}$ frame can be formally denoted as 
\begin{align}
\mathcal{V}_b = [\omega_b, \upsilon_b]^\intercal \in \mathbb{R}^6
\end{align}
in which $\omega_b$ and $\upsilon_b$ express the angular velocity and linear velocity of frame $\{b\}$ represented in the body frame, respectively. However, since motions of body frames will be different even if they lie on the same rigid body, we must transform each body twist into a spatial twist $\mathcal{V}_s = [\omega_s, \upsilon_s]^\intercal \in \mathbb{R}^6$, represented in a common space frame \{s\}.

For two frames, $\{b\}$ and $\{s\}$, where $\{b\}$ is a moving body frame and $\{s\}$ is a fixed space frame, let $T_{sb}$ be the transformation matrix from $\{s\}$ to $\{b\}$ and $\dot T_{sb}$ be the time derivative of $T_{sb}$. 

Conveniently, $T_{sb}$ and $\dot T_{sb}$ have the following relationship 
\begin{align}
\begin{split}
\label{twistDerivation}
\dot T_{sb}T_{sb}^{-1} 
&= \begin{bmatrix} \dot R & \dot p \\ 0 & 0 \\ \end{bmatrix} \begin{bmatrix} R^T & -R^Tp \\ 0 & 1 \\ \end{bmatrix} \\ 
&= \begin{bmatrix} \dot RR^T & \dot p - \dot RR^Tp \\ 0 & 0 \\ \end{bmatrix} = \begin{bmatrix} [\omega_s] & \upsilon_s \\ 0 & 0 \\ \end{bmatrix}
\end{split}
\end{align}
where symbols $R$, $\dot R$, $p$, and $\dot p$ have subscript $sb$ dropped to reduce clutter. $[\omega_s]_{3\times 3}$ is the skew-symmetric representation of $\omega_s$. By this relationship, we are able to calculate the spatial twists $\mathcal{V}_s$ of each body frame in the space frame $\{s\}$. Furthermore, writing $\upsilon_s$ as 
\begin{align}
\upsilon_s 
&= \dot p - \omega_s \times p = \dot p + \omega_s \times (-p)
\end{align}
allows us to infer the physical meaning of $\upsilon_s$. Intuitively, if we imagine a moving rigid body to be infinitely large, $\upsilon_s$ is the instantaneous linear velocity of the point on this body currently at the space frame's origin expressed in the space frame \cite{ModernRobotics}. Fig.~\ref{SampledFrames} illustrates this concept that spatial velocity vector $\upsilon_{\{s\}}^{\{a_1\}/\{a_2\}}$ is the same for both body frames $\{a_1\}$ and $\{a_2\}$ despite different body velocities $\upsilon_{\{a_1\}}$ and $\upsilon_{\{a_2\}}$ (see closeup in Fig.~\ref{SampledFrames}). The same is shown
for spatial velocity vector $\upsilon_{\{s\}}^{\{b_1\}/\{b_2\}}$, which corresponds to body frames $\{b_1\}$ and $\{b_2\}$. It should be noted that spatial velocity vectors $\upsilon_{\{s\}}^{\{a_1\}/\{a_2\}}$ and $\upsilon_{\{s\}}^{\{b_1\}/\{b_2\}}$ are not the same.

Given the transformation $T_{sb}$ and its time derivative $\dot T_{sb}$ between a space frame $\{s\}$ and a body frame $\{b\}$, the instantaneous motions of body frames attached to the same rigid body can be represented as the same spatial twist $\mathcal{V}_s$, regardless of relative body frame motions. This intrinsic characteristic of rigid body motions allows for the distinction of rigid bodies within a scene, so long as their motions are not the same \cite{ModernRobotics}. 

We call this aforementioned spatial twist, $\mathcal{V}_s$, the \emph{Body Frame-Invariant Feature} (BFIF). We denote this feature using the same notation as spatial twist, $\mathcal{V}_s \in \mathbb{R}^6$.

\section{Robot Interactive Object Segmentation}

In Alg.~\ref{RISeg}, we introduced a general interactive perception framework, which included 2 major components: action selection and mask correction. In this section, we will demonstrate how action selection is derived from an uncertainty heatmap produced by a static segmentation model: Mean Shift Mask Transformer for UOIS (MSMFormer) \cite{MSMFormer}, as well as how segmentation masks are corrected based on BFIF grouping derived from an optical flow frame tracking model: Recurrent All Pairs Field Transforms for Optical Flow (RAFT) \cite{RAFT}.

\subsection{Action Selection}

As detailed in Alg.~\ref{FindAction}, we introduce a heuristic-based approach to finding minimal, non-disruptive robot actions, which ensures that the integrity of a given environment is not jeopardized by our interactive perception method. Given an RGB-D image $I_t \in [0, 255]^{H \times W \times 3} \times \mathbb{R}_{+}^{H \times W}$, $\Call{MSMFormer}{\cdot}$ returns segmentation mask $L_t \in \mathbb{Z}_{+}^{H \times W}$ and uncertainty heatmap $U_t \in [0, 255]^{H \times W}$. $U_t$ gives pixel-wise confidence values for each pixel belonging to an object, where pixels with larger values are more likely to belong to an object. In lines 2 and 3 of Alg.~\ref{FindAction}, we use heatmap $U_t$ to identify cluster centers for pixels we are ``certain'' (superscript $c$) to be part of an object, $\{C^c_m\}_{m=1}^{M}$, where $C_{m}^c \in [0, H] \times [0, W]$. Heatmap $U_t$ is also used to identify cluster centers for pixels we are ``uncertain'' (superscript $u$) to be part of an object, $\{C^u_n\}_{n=1}^{N}$, where $C_{n}^u \in [0, H] \times [0, W]$. Threshold values $\ell_{u}$ and $\ell_{l}$ are used to perform this clustering, where $\ell_{u} > \ell_{l}$ and $N \gg M$. Formally, cluster centers $\{C^c_m\}$ are derived from k-means clustering on pixels $(i,j)$ in uncertainty heatmap $U_t$ where $U^{i,j}_t \geq \ell_{u}$ such that pixels $(i,j)$ of $U_t$ are pixels we are ``certain'' belong to an object. Similarly, cluster centers \{$C^u_n$\} are derived from k-means clustering on pixels $(i,j)$ in uncertainty heatmap $U_t$ where $\ell_{l} \leq U^{i,j}_t < \ell_{u}$ such that pixels $(i,j)$ of $U_t$ are pixels we are ``uncertain'' of belonging to an object or not. The number of clusters $M$ and $N$ for cluster centers $\{C^c_m\}_{m=1}^{M}$ and $\{C^u_n\}_{n=1}^{N}$ are derived via the elbow method. Threshold values $\ell_{u}$ and $\ell_{l}$ are identified via experimentation.

\begin{figure}[htbp]
\vspace{-5pt}
\begin{algorithm}[H]
\footnotesize
\caption{FindAction}\label{FindAction}
\textbf{Input:} $I_{t}$ \\
\textbf{Output:} $a_{t}$
\begin{algorithmic}[1]
\State $L_{t}, U_{t} \leftarrow \Call{MSMFormer}{I_{t}}$
\State $\{C^c_m\}_{m=1}^{M} \leftarrow \Call{KMeans}{U^{i,j}_{t} \in U_{t}: \ell_{u} \leq U^{i,j}_{t}}$
\State $\{C^u_n\}_{n=1}^{N} \leftarrow \Call{KMeans}{U^{i,j}_{t} \in U_{t}: \ell_{l} \leq U^{i,j}_{t} < \ell_{u}}$
\State $(i^*,j^*) \leftarrow \underset{(i,j) \in \{1,...,M\}}{\arg\min} \|C^c_{i}-C^c_{j}\|$ \par
\hspace{1pt}s.t. $i \neq j$, \par
\hspace{1pt}\phantom{s.t.} $\|C^c_{i}-C^c_{j}\| \leq d_{a}$, \par
\hspace{1pt}\phantom{s.t.} $\underset{n\in \{1,...,N\}}{\min} \Call{Dist}{C^u_{n},\overline{C^c_{i}C^c_{j}}\}}\leq d_{b}$ 
\If{$(i^*,j^*)$ exists}
\State $\{P_{i^*}\} \leftarrow \Call{Boundary}{C^c_{i^*}}$
\State $P^{*} \leftarrow \Call{Rand}{\{P_{i} \in \{P_{i^*}\}: \overline{P_{i}C^c_{i^*}} \perp \overline{C^c_{i^*}C^c_{j^*}} \}}$
\State $a_{t} \leftarrow (P^{*}, \overrightarrow{P^{*}C^c_{i^*}}, d_{push})$
\State \textbf{return} $a_{t}$
\Else
\State \textbf{return} \textit{null}
\EndIf

\end{algorithmic}
\end{algorithm}
\vspace{-20pt}
\end{figure}

\begin{figure}[t]
\includegraphics[width=\columnwidth]{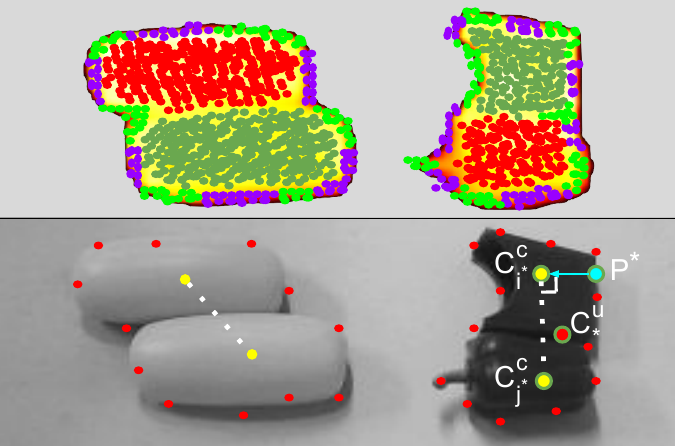}
\centering
\caption{Visualization of FindAction($\cdot$). [Top] ``Certain'' clusters shown in red and dark green. ``Uncertain'' clusters shown in purple and light green. [Bottom] ``Certain'' cluster centers ($C^c_m$) are shown in yellow. White, dashed line segments connect ``certain'' cluster centers ($\overline{C^c_{i}C^c_{j}}$). ``Uncertain'' cluster centers ($C^u_n$) are shown in red. Action $a_t$, defined by chosen push point $P^*$ and direction $\overrightarrow{P^{*}C^c_{i^*}}$, is shown in blue. ``Uncertain'' cluster center $C^u_*$ is used to choose $C^c_{i^*}$ and $C^c_{j^*}$ due to having minimum distance to $\overline{C^c_{i^*}C^c_{j^*}}$.}
\label{findActionFig}
\vspace{-15pt}
\end{figure}

Fig.~\ref{findActionFig} shows how a specific robot action is selected after obtaining the ``certain'' and ``uncertain'' clusters from uncertainty heatmap $U_t$. With cluster centers \{$C^c_m$\} and \{$C^u_n$\}, we must select two ``certain'' clusters for which we wish to interact with and ``learn'' more about. In line 4 of Alg.~\ref{FindAction}, we describe consideration of all pairs $(i,j)$ of cluster centers in \{$C^c_m$\} where $i \neq j$ and the distance between $C^c_i$ and $C^c_j$ is less than some distance $d_{a}$. A distance constraint $d_{a}$ is necessary to avoid selecting objects far from one another. For each $(C^c_i$, $C^c_j)$ pair under consideration, we construct a line segment connecting the cluster center pair, and select the pair of interest ($C^c_{i^*}$, $C^c_{j^*}$) for which an ``uncertain'' cluster center $C^u_n$ is closest to. The distance between ``uncertain'' cluster center $C^u_n$ and line segment $\overline{C^c_{i}C^c_{j}}$ must be at most $d_b$. Otherwise, we can say that there is not enough uncertainty to explore those clusters. If no ``certain'' cluster centers  $C^c_{i^*}$ and $C^c_{j^*}$ exist to satisfy these constraints, then no qualifying action $a_t$ exists, and a \textit{null} action will be returned.

Once ``certain'' clusters $C^c_{i^*}$ and $C^c_{j^*}$ are heuristically identified, we can generate a specific action, $a_t$, by first identifying a push point and then a direction (see Fig~\ref{findActionFig}). Push point $P^*$ is chosen by first obtaining pixels \{$P_{i^*}$\} from the cluster boundary of cluster center $C^c_{i^*}$ via $\Call{Boundary}{\cdot}$. Then, a point $P^*$ that forms a line segment $\overline{P^*C^c_{i^*}}$ perpendicular to line segment $\overline{C^c_{i^*}C^c_{j^*}}$ is chosen at random via line 7 of Alg.~\ref{FindAction}. Action $a_t$ is now defined as a push from point $P^*$ in direction $\overrightarrow{P^{*}C^c_{i^*}}$ for short constant distance $d_{push}$. This push point and direction is chosen to reduce the possibility of clusters $C^c_{i^*}$ and $C^c_{j^*}$ moving in the same direction. A small distance $d_{push}$ is selected to reduce disruption of the given scene as a result of action $a_t$. Once action $a_t$ is transformed from the image space to the robot workspace via the camera matrix, $a_t$ is executed, and new image $I_{t+1}$ and segmentation mask $L_{t+1}$ are captured. 

\begin{figure}[t]
\vspace{-5pt}
\begin{algorithm}[H]
\footnotesize
\caption{UpdateMask}\label{UpdateMask}
\textbf{Input:} $I_{t}, I_{t+1}, \hat{L}_{t}, L_{t+1}$ \\
\textbf{Output:} $\hat{L}_{t+1}$
\begin{algorithmic}[1]
\State $O_{t} \leftarrow \Call{RAFT}{I_{t}, I_{t+1}}$ \hfill \Comment{Optical Flow}
\State $\{F_{t}^{i}\}, \{F_{t+1}^{i}\} \leftarrow \Call{CreateFrames}{\hat{L}_{t}, O_{t}}$
\State $\{\mathcal{V}_{t}^{i}\} \leftarrow \Call{CalcBFIFs}{\{F_{t}^{i}\}, \{F_{t+1}^{i}\}}$
\State $FG_{t} \leftarrow \Call{GroupBFIFs}{\{\mathcal{V}_{t}^{i}\}, \hat{L}_{t}}$
\State $\hat{L}_{t+1} \leftarrow \Call{CorrectMask}{FG_{t}, \hat{L}_{t}, L_{t+1}, O_{t}}$
\State \textbf{return} $\hat{L}_{t+1}$
\end{algorithmic}
\end{algorithm}
\vspace{-20pt}
\end{figure}

\begin{figure*}[h]
\includegraphics[width=1.5\columnwidth]{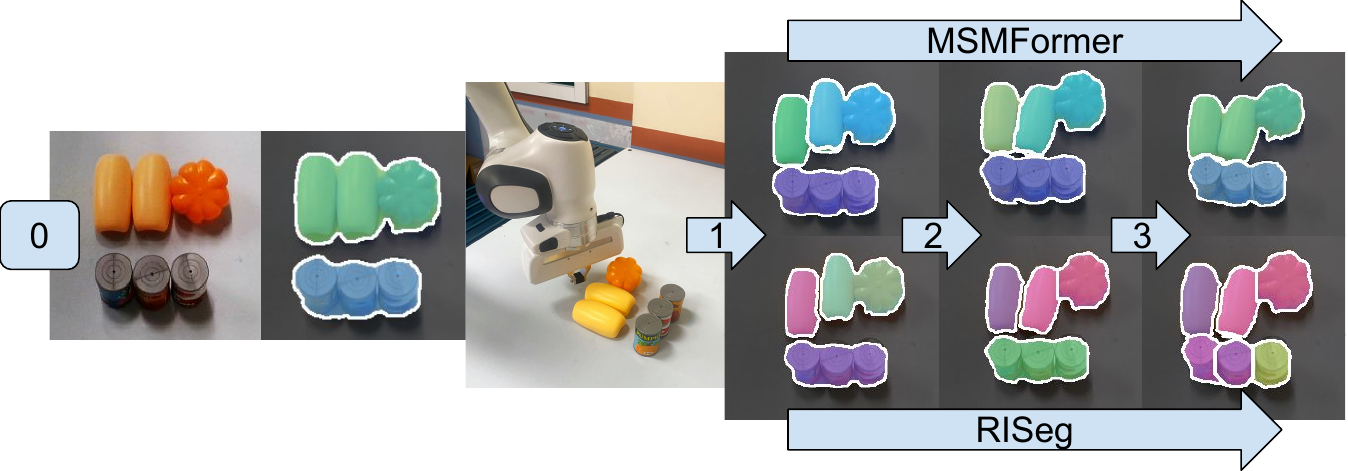}
\centering
\caption{RISeg and MSMFormer segmentations of a cluttered tabletop scene throughout the interactive perception pipeline. The scene's initial state is shown after label ``0''. Scene configurations and segmentation masks after push numbers 1, 2, and 3 follow the corresponding arrows. Pushes are minimal and are always less than 2cm.}
\label{seg-pipeline}
\vspace{-12pt}
\end{figure*}

\subsection{Segmentation Mask Correction}

\subsubsection{Sample Body Frames and Compute BFIFs}

Since a main motivation of our method is to improve segmentation through non-disruptive interactions, $I_t$ and $I_{t+1}$ will be visually very similar to one another. Therefore, $L_{t+1}$ is still likely to have similar segmentation inaccuracies as $L_{t}$, such as under segmentation. In Alg.~\ref{UpdateMask}, we describe how even without object singulation in $I_{t+1}$, we are able to produce a more accurate, refined segmentation mask $\hat{L}_{t+1}$ for the current scene state.

To track motions caused by robot interactions, we use an optical flow model $\Call{RAFT}{\cdot}$, which given input images $I_t$ and $I_{t+1}$, outputs a gradient map of pixel motions $O_{t} \in (\mathbb{R}, \mathbb{R})^{H \times W}$. To compute the BFIFs of objects between scene images $I_t$ and $I_{t+1}$, we must create body frames attached to rigid bodies in $I_t$ and track their motion through to $I_{t+1}$. Creating such body frames via $\Call{CreateFrames}{\cdot}$ involves 3 steps. First, we must sample $n$ random pixels which belong to an object from $\hat{L}^{i,j}_t \in \hat{L}_t$ where $\hat{L}^{i,j}_t \neq 0$. Then, we pick triplets of pixels among those sampled to create frames. Each triplet of pixels selected to create each frame should not be collinear and should have a maximum distance between them of $d_{c}$. A frame can then be created by picking one point to be the origin and using the other two points to find directions for each axis. The \textit{z}-axis is perpendicular to the plane formed by the triplet of sampled points, the \textit{x}-axis is formed by connecting the origin with one of the other two points, and the \textit{y}-axis is perpendicular to the \textit{x} and \textit{z} axes. Finally, $O_t$ is used to track the sampled frames between $I_t$ and $I_{t+1}$. 

With a set of body frames $\{F_{t}^{i}\}$ from $I_t$ and a corresponding set of body frames $\{F_{t+1}^{i}\}$ from $I_{t+1}$, we can compute a set of BFIFs $\{\mathcal{V}_{t}^{i}\}$ represented in the space frame $\{s\}$, as described by Equation~\ref{twistDerivation}. In $\Call{CalcBFIFs}{\cdot}$, transformation matrices $T_{sb}$ and time derivative $\dot T_{sb}$ are derived from each body frame pair ($F_{t}^{i}, F_{t+1}^{i}$) and the space frame $\{s\}$, which then allows for the computation of each BFIF $\mathcal{V}_{t}^{i}$. In this work, the space frame $\{s\}$ is selected to be the camera frame for simplicity. Remember that BFIFs in $\{\mathcal{V}_{t}^{i}\}$ will theoretically be equal if they belong to body frames on the same rigid body. However, due to noise in optical flow $O_t$, computed BFIFs for each body frame have slight inaccuracies. Therefore, we choose to filter out the noise by using a statistical model to group BFIFs with one another.

\subsubsection{BFIF Grouping}

$\Call{GroupBFIFs}{\cdot}$ aims to identify if two frames lie on the same rigid body given the difference in their corresponding BFIFs by pairwise BFIF comparisons via Bayesian Inference. We formulate this statistical model as 
\begin{align}
\footnotesize
\overbrace{P(\textrm{hypothesis} \vert \textrm{data})}^{\textrm{posterior}} = \frac{\overbrace{P(\textrm{data} \vert \textrm{hypothesis})}^{\textrm{likelihood}} \overbrace{P(\textrm{hypothesis})}^{\textrm{prior}}}{\underbrace{P(\textrm{data})}_{\textrm{evidence}}}
\end{align}
where \textit{hypothesis} is defined as two body frames belonging to the same rigid body/object and \textit{data} is defined as the difference in BFIFs (spatial twists) represented in the space frame for those two body frames. Mathematically, for some body frame $\{b_i\}$ with origin $q_i$ and BFIF $\mathcal{V}_t^i$ and some other body frame $\{b_j\}$ with origin $q_j$ and BFIF $\mathcal{V}_t^j$, \textit{hypothesis} can be written as an indicator function
\begin{align}
X_{i,j} = \mathds{1}(\hat{L}_t(q_i) = \hat{L}_t(q_j)) \in \{0,1\}
\end{align}
and \textit{data} can be written as 
\begin{align}
Y_{i,j} = \textrm{diff}(\mathcal{V}_t^i,\mathcal{V}_t^j)
\end{align} where $i \neq j$.
Given the above formalizations of the data and hypothesis, we use Kernel Density Estimation (KDE) \cite{Weglarczyk2018KDE} to estimate the \textit{Posterior} and group pairs of BFIFs. The unions of intersecting grouped body frame pairs are used to form a set of sets of frames $FG_t$, where each inner set contains frames identified to have similar BFIFs. Frame groups $FG_t$ can be expanded as $FG_t = \{fg_0, fg_1, fg_2, \ldots\}$, where each set of body frames $fg_i$ is comprised of body frames identified to have the same BFIF.

\subsubsection{Segmentation Mask Correction}
Once we have identified body frame groups $FG_t$, we can correct segmentation inaccuracies in $L_{t+1}$, via line 5 of Alg.~\ref{UpdateMask} $\Call{CorrectMask}{\cdot}$, and return $\hat{L}_{t+1}$. To do so, we first project $\hat{L}_{t}$ object segmentations onto corresponding objects in $\hat{L}_{t+1}$, and then use the grouped body frames $FG_t$ with similar BFIFs to correct $\hat{L}_{t+1}$.

By using the most recent RISeg segmentation mask $\hat{L}_t$ as an accumulation of previous mask corrections, we first bring the current RISeg mask $\hat{L}_{t+1}$ to the same level of segmentation accuracy as $\hat{L}_t$, which will reflect the information gained from all previous interactions $a_{t-1}, a_{t-2}, \ldots$. Optical flow $O_t$ is used to map each labeled pixel in $\hat{L}_{t}$ to the corresponding pixel in $\hat{L}_{t+1}$. Once $\hat{L}_{t+1}$ reflects the segmentation masks of $\hat{L}_{t}$ by using the aforementioned mappings, we can use the grouped body frames $FG_t$ to correct $\hat{L}_{t+1}$, which will reflect the information gained from interaction $a_t$.

Each set $fg_i \in FG_t$ represents a group of body frames identified to have the same BFIF. Therefore, each body frame in set $fg_i$ should be segmented as part of the same object with object ID $\ell_i$, along with similarly moving neighboring points. For each body frame in $fg_i$, we reassign its corresponding pixel in $\hat{L}_{t+1}$ to $\ell_i$. These initial $\hat{L}_{t+1}$ pixel reassignments act as seed points for object $\ell_i$ since the number of sampled body frames $n$ in line 2 of Alg~\ref{UpdateMask} is very small relative to total number of pixels $H\times W$. Once the seed points have been set for new label $\ell_i \in \hat{L}_{t+1}$, Breadth First Search is used to assign object ID $\ell_i$ to pixels that move with similar gradient compared to pixels already reassigned to object ID $\ell_i$, starting from the seed points and expanding outwards.

It should be noted that BFIFs must first be used to find seed points rather than directly using the optical flow gradient because BFIFs are body frame-invariant. After each set $fg_i \in FG_t$ has corrected the corresponding pixels, we return new segmentation mask $\hat{L}_{t+1}$.

\section{Experiment}

In this section, we demonstrate that the proposed RISeg is an effective framework for Interactive Perception of unseen objects in cluttered environments by comparison with state-of-the-art method MSMFormer \cite{MSMFormer}. Our experiments showcase that segmentation can be drastically improved by using small, non-disruptive pushes and tracking BFIFs. Fig~\ref{seg-pipeline} shows a qualitative comparison of segmentation results between MSMFormer and RISeg.

\subsection{Implementation and Dataset}

\textbf{Experiment set up.} The RISeg Interactive Perception framework uses a Franka Emika 7dof robot \cite{FrankaEmika} to perform robot interactions with the scene and an Intel Realsense D415 RGB-D Camera \cite{D415} to capture real-time visual data. Experiment objects are placed on a flat, white tabletop and come from a set of play food toys for kids due to similarity in shape and color to one another. These objects are particularly difficult to segment in cluttered environments. The D415 camera is placed approximately 60cm above the tabletop with an angle of 15 degrees to the vertical axis. 

\textbf{RISeg implementation.} In lines 2 and 3 of Alg.~\ref{FindAction}, we describe two threshold values, $\ell_u$ and $\ell_l$, for k-means clustering on uncertainty heatmap $U_t$. In our work, $\ell_u = 150$ and $\ell_l = 120$. Furthermore, line 4 of Alg.~\ref{FindAction} describes a maximum distance threshold for considering ``certain'' cluster pairs, $d_{a}$, which we define to be 10cm. Finally, line 7 of Alg.~\ref{FindAction} describes a constant $d_{push}$ for the distance of each robot action $a_t$, which is defined to be 2cm. 

\textbf{Experiment dataset.} Because there is no standard interactive perception dataset, we evaluate our proposed pipeline by creating 23 tabletop scenes in which 4-6 objects are placed in close proximity to one another, often touching. For each scene, 2-3 robot interactions, as automatically determined by Alg.~\ref{FindAction}, are completed. This results in roughly 78 total images across all scenes and interactions to evaluate segmentation on. For each of these images, we create ground truth masks by manual annotations.

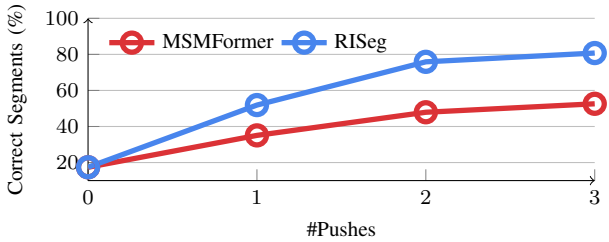
\begin{figure}[t]
%
%
\definecolor{mycolor1}{rgb}{0.85882,0.19608,0.21176}%
\definecolor{mycolor2}{rgb}{0.28235,0.52157,0.92941}%
\definecolor{mycolor3}{rgb}{0.23529,0.72941,0.32941}%
\definecolor{mycolor4}{rgb}{0.95686,0.76078,0.05098}%
\definecolor{mycolor5}{rgb}{1.00000,0.54902,0.00000}%
\begin{tikzpicture}

\begin{axis}[%
width=0.78\columnwidth,
height=0.25\columnwidth,
scale only axis,
xmin=0,
xmax=3,
xlabel={\#Pushes},
xtick={0,1,...,3},
ymin=10,
ymax=100,
ylabel={Correct Segments (\%)},
axis background/.style={fill=white},
title style={font=\bfseries},
ymajorgrids,
legend style={at={(0.02,0.85)}, anchor=west, legend cell align=left, align=left, draw=white!15!black, fill=none, draw=none, legend columns = 2}
]
\addplot [color=mycolor1, line width=2.0pt, mark size=3.6pt, mark=o, mark options={solid, mycolor1}]
  table[row sep=crcr]{%
0	17.39130435\\
1	35.07246377\\
2	47.89855072\\
3	52.59259259\\
};
\addlegendentry{MSMFormer}

\addplot [color=mycolor2, line width=2.0pt, mark size=3.6pt, mark=o, mark options={solid, mycolor2}]
  table[row sep=crcr]{%
0	17.39130435\\
1	51.88405797\\
2	75.86956522\\
3	80.74074074\\
};
\addlegendentry{RISeg}

\end{axis}
\end{tikzpicture}%
\centering
\caption{Percentage of objects correctly segmented as measured by the Overlap F-measure $\geq 75\%$.}
\label{graph}
\vspace{-12pt}
\end{figure}
 
\subsection{Evaluation Metrics}

For each scene, we evaluate the segmentation accuracy at each scene configuration by comparing results between MSMFormer and RISeg. Scene configurations include initial (push 0), after push 1, after push 2, and after push 3. 

We evaluate the object segmentation performance using precision, recall and F-measure \cite{Xiang2021UOIS, Xie2019Both}. For each metric, we compute values between all pairs of predicted objects and ground truth objects. Then, the Hungarian method and pairwise F-measure are used to match predictions with the ground truth. Precision, recall, and F-measure can therefore be defined as $P=\frac{\sum_i \vert c_i \cap g(c_i)\vert}{\sum_i \vert c_i \vert}$, $R=\frac{\sum_i \vert c_i \cap g(c_i)\vert}{\sum_j \vert g_j \vert}$, $F=\frac{2PR}{P+R}$, where $c_i$ denotes the segmentation mask of predicted object $i$ and $g(c_i)$ and $g_j$ denote the segmentation mask of the matched ground truth object of $c_i$ and the ground truth object $j$. 

In Table~\ref{seg-table}, we show these 3 metrics under the ``Overlap'' column since these true positives can be viewed as the overlap between prediction and ground truth segmentations. Additionally, boundary P/R/F metrics are used to evaluate how sharp predicted boundaries are in comparison to ground truth boundaries. True positives for boundaries are counted by the pixel overlap of the two boundaries. Furthermore, Fig.~\ref{graph} shows the percentage of objects segmented with a high accuracy throughout scene configurations, which is the percentage of segmented objects with Overlap F-measure $\geq 75\%$. 

\begin{table}
\centering
\footnotesize
\captionsetup{labelformat=empty} 
\setlength{\tabcolsep}{5pt}
\begin{tabular}{|c|c|c c c|c c c|}           
\hline 
\multirow{2}{*}{Method} &\multirow{2}{*}{Push \#}    & \multicolumn{3}{c|}{Overlap}  & \multicolumn{3}{c|}{Boundary}                   \\   
 &  &{P} & {R} & {F}   &{P} & {R} & {F}           \\   \hline 
\multirow{4}{*}{MSMFormer \cite{MSMFormer}} 
 & 0 & 53.7 & 55.4 & 52.3 & 44.6 & 50.6 & 40.0                \\
 & 1 & 66.6 & 62.4 & 64.3 & 62.1 & 52.4 & 56.8                   \\   
 & 2 & 72.8 & 68.6 & 70.5 & 69.0 & 61.1 & 64.7                    \\   
 & 3 & 73.2 & 67.6 & 70.1 & 70.0 & 62.5 & 65.9              \\   \hline
 \multirow{4}{*}{RISeg}
 & 0 & 53.7 & 55.4 & 52.3 & 44.6 & 50.6 & 40.0                   \\
 & 1 & 74.1 & 69.6 & 71.6 & 69.0 & 61.5 & 64.9     \\     
 & 2 & 85.8 & 81.1 & 83.3 & 79.4 & 76.0 & 77.6                \\    
 & 3 & \textbf{88.1} & \textbf{79.6} & \textbf{83.3} & \textbf{82.4} & \textbf{77.4} & \textbf{79.6}                \\   
 \hline
 \end{tabular}
 \caption{\label{seg-table}Table 1. Segmentation results of MSMFormer and RISeg across scene configurations resulting from robot actions.}
 \vspace{-12pt}
 \end{table}

\subsection{Discussion of Results}

In Table~\ref{seg-table} and Fig.~\ref{graph}, we compare segmentation results of our RISeg method with state-of-the-art UOIS model MSMFormer. Push 0 indicates the scene's initial configuration, in which both methods have the same segmentation results because RISeg uses MSMFormer for base segmentation masks. Each push number indicates average segmentation statistics across all scenes after that numbered interaction has been completed, regardless of total number of pushes for each individual scene. Initially, both methods accurately segment less than 20\% of total objects. With each robot-scene interaction, both methods see object segmentation accuracy increases for all metrics, though to different degrees. On average, MSMFormer object segmentation accuracy increases with each interaction because interactions are more likely to result in some object singulation than not. However, RISeg object segmentation accuracy increases drastically faster and sees a higher peak when compared to MSMFormer because analysis of BFIFs results in robust segmentations even with minimal object displacements and no object singulation. After all robot interactions, RISeg is able to accurately segment 80.7\% of objects in the scene's end configuration while MSMFormer is still only able to segment 52.5\% of objects. Overlap and Boundary P/R/F metrics also increase with each robot interaction. Overlap precision metrics peak after interaction number 3 is completed, with 88.1\% for RISeg and 73.2\% for MSMFormer. 

\section{Conclusion}
In this work, we proposed an Interactive Perception pipeline, RISeg, which uses minimal non-disruptive interactions to segment a scene by tracking designed Body Frame-Invariant Features (BFIFs). This designed feature uses the insight that two body frames attached to the same rigid body experiencing different rotations and translations in space will have the same spatial twist observed by any fixed world frame. We then demonstrated the effectiveness of RISeg in segmenting real-world tabletop scenes of cluttered difficult-to-segment objects. In future work, we plan to explore video-based frame tracking to analyze object motions throughout a single interaction rather than only start and end states. 

\newpage

\bibliographystyle{IEEEtran}
\bibliography{conference_101719}

\end{document}